\documentclass[11pt]{article}

\usepackage[margin=1in]{geometry}
\usepackage{amsmath,amssymb}
\usepackage{graphicx}
\usepackage{booktabs}
\usepackage{multirow}
\usepackage{hyperref}
\usepackage[numbers]{natbib}
\usepackage{tikz}
\usetikzlibrary{shapes.geometric, arrows.meta, positioning, fit}
\usepackage{caption}
\usepackage{subcaption}
\usepackage{enumitem}
\usepackage{xcolor}
\usepackage{float}

\hypersetup{
    colorlinks=true,
    linkcolor=blue,
    citecolor=blue,
    urlcolor=blue
}

\title{Evaluating the Effect of Frame Rate in Sequence-Based Classification of Autism-Related Self-Stimulatory Hand Idiosyncrasies}

\author{
Raunak Mondal$^{1}$ and Peter Washington$^{2}$ \\[6pt]
\small{$^{1}$School of Computer Science, Carnegie Mellon University, Pittsburgh, PA, USA} \\
\small{$^{2}$Department of Information and Computer Sciences,} \\
\small{University of Hawai`i at M\={a}noa, Honolulu, HI, USA}
}

\date{}

\begin{document}

\maketitle

\begin{abstract}
Autism spectrum disorder (ASD) affects over 75 million individuals worldwide, yet scalable computational methods for remote behavioral screening remain limited. This study addresses two complementary challenges in automated detection of autism-related self-stimulatory behaviors from video: (1) identifying the optimal sequence-based neural network architecture and temporal sampling rate, and (2) characterizing data augmentation strategies for training on small behavioral datasets. For the first objective, long short-term memory (LSTM) and gated recurrent unit (GRU) models were trained on pose-derived features from the Self-Stimulatory Behavior Diagnosis (SSBD) dataset at frame sampling intervals of 1, 5, 15, 30, 45, and 90 frames. Both architectures exceeded prior convolutional neural network (CNN) baselines (62--76\% accuracy), with peak accuracies of 97.5\% (LSTM) and 98.75\% (GRU) at a sampling interval of every 15 frames. For the second objective, ten data augmentation strategies were applied to an I3D transfer learning pipeline, with an ablation study quantifying the marginal contribution of each technique. Horizontal flip achieved the highest standalone accuracy (48.78\%), while exclusion of upsampling from the augmentation pipeline produced the largest performance degradation, indicating its necessity for augmentation applied to complex behavioral video. A personalized machine learning approach, in which per-subject models were trained and tested on temporally split segments of each video, produced consistent predictions (mean loss 1.84, SD 0.79). These results provide practitioners with concrete guidance on architecture selection, sampling rate, and augmentation strategy for video-based behavioral classification in data-scarce clinical domains.
\end{abstract}

\section{Introduction}

Autism spectrum disorder (ASD) is a neurodevelopmental condition characterized by differences in social communication and the presence of restricted, repetitive behaviors~\cite{american2013diagnostic}. Current estimates place worldwide prevalence at approximately 75 million individuals, with reported rates in the United States increasing 241\% since 2000~\cite{taca2021prevalence}. Despite this high prevalence, diagnosis remains labor-intensive: clinicians rely on structured behavioral observation instruments such as the Autism Diagnostic Observation Schedule (ADOS-2), which require trained administrators and multi-visit assessment pipelines~\cite{lord2012autism}. These constraints create bottlenecks that delay diagnosis, particularly in low-resource settings where specialist access is limited.

Automated recognition of self-stimulatory (``stimming'') behaviors from video offers one path toward scalable screening. Behaviors such as hand flapping, head banging, and spinning are among the motor stereotypies commonly observed in ASD and can in principle be detected from consumer-grade video recordings~\cite{rajagopalan2013ssbd}. However, computational approaches face two interrelated challenges. First, publicly available datasets of labeled stimming behaviors are small. The Self-Stimulatory Behavior Diagnosis (SSBD) dataset~\cite{rajagopalan2013ssbd}, one of the few purpose-built resources, contains only 75 videos across three behavior categories. Second, prior work has relied heavily on convolutional neural network (CNN) models, which achieve state-of-the-art accuracies of only 62--76\% on these tasks~\cite{khodatars2021deep}.

Sequential architectures such as long short-term memory (LSTM) networks and gated recurrent units (GRUs) are well-suited to temporal pattern recognition and have shown strong performance in adjacent activity recognition domains~\cite{greff2017lstm, cho2014gru}. Yet their application to autism-specific behavioral classification remains underexplored~\cite{jacob2022algorithmic}. An additional open question concerns the temporal granularity at which input features should be sampled: extracting pose or motion features at every frame may introduce redundancy, while overly aggressive downsampling risks discarding diagnostically relevant temporal dynamics.

In parallel, the small size of behavioral datasets motivates investigation of strategies to improve model robustness without additional data collection. Transfer learning from large-scale activity recognition models such as Inflated 3D ConvNet (I3D)~\cite{carreira2017quo} can provide useful feature representations, while data augmentation techniques can synthetically expand the effective training set. However, the relative effectiveness of different augmentation strategies for clinical behavioral video has not been systematically characterized.

This paper presents two complementary sets of experiments conducted on the SSBD dataset. The first set evaluates LSTM and GRU architectures across six temporal sampling rates to identify the architecture and sampling configuration that maximize classification accuracy. The second set applies ten data augmentation methods within an I3D transfer learning framework, accompanied by an ablation study and a personalized machine learning protocol. Together, these experiments provide concrete, actionable guidance for researchers developing video-based tools for autism behavioral screening.

The contributions of this work are as follows:
\begin{enumerate}[nosep]
    \item A systematic comparison of LSTM and GRU architectures at six frame sampling intervals for self-stimulatory behavior classification, demonstrating that both architectures surpass prior CNN baselines and that a sampling interval of 15 frames yields peak performance.
    \item A characterization of ten data augmentation strategies for I3D-based transfer learning on small behavioral video datasets, including an ablation study identifying upsampling as the single most critical augmentation component.
    \item Evaluation of a personalized machine learning approach in which per-subject models are trained on temporally partitioned segments of individual videos.
\end{enumerate}

\section{Related Work}

\subsection{Computational Approaches to Autism Behavioral Detection}

Machine learning methods for autism detection span a range of modalities and architectures. Shahamiri and Thabtah~\cite{shahamiri2020autism} developed Autism AI, a screening system combining multiple machine learning classifiers to improve detection accuracy relative to single-model approaches. Abbas et al.~\cite{abbas2020multimodular} proposed a multi-modular framework integrating video, questionnaire, and sensor data for streamlined diagnosis in young children, demonstrating that combining data sources can improve classification over unimodal approaches. Ghosh et al.~\cite{ghosh2021ai_iot} surveyed the intersection of AI and Internet-of-Things technologies for ASD screening and management, noting the potential of wearable and ambient sensors to capture behavioral data in naturalistic settings. Wall et al.~\cite{wall2012ai_autism} applied machine learning to shorten the behavioral diagnosis process, showing that a subset of ADOS items could achieve comparable diagnostic accuracy to the full instrument. These studies collectively demonstrate growing interest in computational autism screening, but the majority rely on hand-crafted features or CNN architectures applied to static frames rather than exploiting the temporal structure of behavioral sequences.

\subsection{Activity Recognition with Deep Learning}

Automated recognition of human activities from video is a well-studied problem in computer vision. Mo et al.~\cite{mo2016human} achieved 81.8\% accuracy on 12-activity classification by feeding CAD-60 skeletal pose sequences into a CNN. Li and Chuah~\cite{li2018rehar} proposed ReHAR, a framework combining CNN and LSTM components for robust activity recognition from sensor data. Carreira and Zisserman~\cite{carreira2017quo} introduced the Inflated 3D ConvNet (I3D), which extends 2D convolutional filters to the temporal dimension using weights pretrained on the Kinetics dataset, achieving strong transfer performance on smaller downstream activity recognition tasks. The I3D architecture is particularly relevant to the present work because it provides a mechanism for transferring learned spatiotemporal representations to data-scarce domains such as autism behavioral analysis.

\subsection{Sequence-Based Models for Temporal Classification}

Recurrent neural networks, and in particular LSTM~\cite{hochreiter1997lstm} and GRU~\cite{cho2014gru} variants, have demonstrated strong performance on sequential classification tasks across domains including speech recognition, natural language processing, and time-series analysis~\cite{greff2017lstm}. In the activity recognition literature, LSTM-based models have been applied to skeleton-based action recognition~\cite{du2015hierarchical} and video captioning~\cite{donahue2015long}. GRUs offer a computationally lighter alternative with fewer parameters, often achieving comparable accuracy at reduced training time~\cite{chung2014empirical}. Despite their suitability for capturing temporal dependencies in behavioral sequences, systematic evaluation of these architectures on autism-specific datasets has been limited.

\subsection{Data Augmentation for Video-Based Models}

Data augmentation is a standard strategy for improving generalization in data-constrained settings. For video and activity recognition, augmentation operates along both spatial and temporal dimensions. Spatial augmentations include horizontal and vertical flips, cropping, and noise injection; temporal augmentations include speed perturbation, temporal elastic deformation, resampling, and frame reordering~\cite{shorten2019survey}. Li et al.~\cite{li2020data_aug_dense} proposed dense joint motion images for augmenting skeleton-based action recognition data, while Luo et al.~\cite{luo2021activitygan} introduced ActivityGAN, a generative adversarial approach for synthesizing sensor-based activity data. Kim et al.~\cite{kim2020timeseries_aug} applied time-series augmentation with deep learning to construction equipment activity recognition, demonstrating improvements in both accuracy and generalization. The effectiveness of these strategies is domain-dependent, and their impact on clinical behavioral video classification has not been previously characterized.

\subsection{Prior Work on the SSBD Dataset}

The Self-Stimulatory Behavior Diagnosis (SSBD) dataset was introduced by Rajagopalan et al.~\cite{rajagopalan2013ssbd}, who established baseline classification results using bag-of-words representations with STIP and HIG/HOF features fed into a support vector machine (SVM), reporting a mean accuracy of 47.1\% across codebook sizes of 500, 1000, and 1500 on the three-class task. In subsequent work~\cite{rajagopalan2014detecting}, the same group achieved 86.6\% binary accuracy on head banging versus spinning and 76.3\% on the three-way classification task. Lakkapragada et al.~\cite{lakkapragada2022classification} applied MobileNetV2 feature extraction to the SSBD hand flapping subset, achieving an F1 score of 84 (precision 89.6, recall 80.4) using five-fold cross-validation across 100 random seeds (500 total folds). The present study builds on these results by evaluating sequential architectures (LSTM, GRU) and transfer learning with data augmentation on the same dataset.

\section{Methods}

\subsection{Dataset}

All experiments used the Self-Stimulatory Behavior Diagnosis (SSBD) dataset~\cite{rajagopalan2013ssbd}, which contains videos of children exhibiting self-stimulatory behaviors categorized as arm flapping, head banging, and spinning. The dataset comprises 75 annotated videos with an average duration of approximately 90 seconds per video. Annotations are provided in XML format, specifying temporal boundaries for each behavior instance. All videos in the dataset were collected from publicly available recordings that parents or caregivers consented to post on the internet~\cite{rajagopalan2013ssbd}, and no independent review board approval was required.

\subsection{Experiment 1: Sequence-Based Architecture Comparison}

\subsubsection{Feature Extraction and Sampling}

For the architecture comparison experiments, a subset of 50 videos containing hand flapping behaviors was used. Each video spans 90 frames. Pose-based geometric features were extracted from each frame using a feature extraction pipeline that computes spatial attributes of detected body keypoints. To investigate the effect of temporal sampling granularity, features were extracted at six sampling intervals: every 1, 5, 15, 30, 45, and 90 frames. At a sampling interval of $k$ frames, features were extracted from frames $\{1, 1+k, 1+2k, \ldots\}$, yielding input sequences of length $\lfloor 90/k \rfloor$.

\subsubsection{Model Architectures}

Two sequential model architectures were evaluated, each consisting of three layers:

\paragraph{LSTM Model.} The first model contained a single LSTM layer followed by a dropout layer (rate = 0.5) and a dense output layer with softmax activation. The LSTM layer processes the input sequence of pose features and produces a fixed-length hidden state representation. The forget, input, and output gates of the LSTM enable selective retention of temporal information across the sequence.

\begin{figure}[H]
\centering
\includegraphics[width=0.85\textwidth]{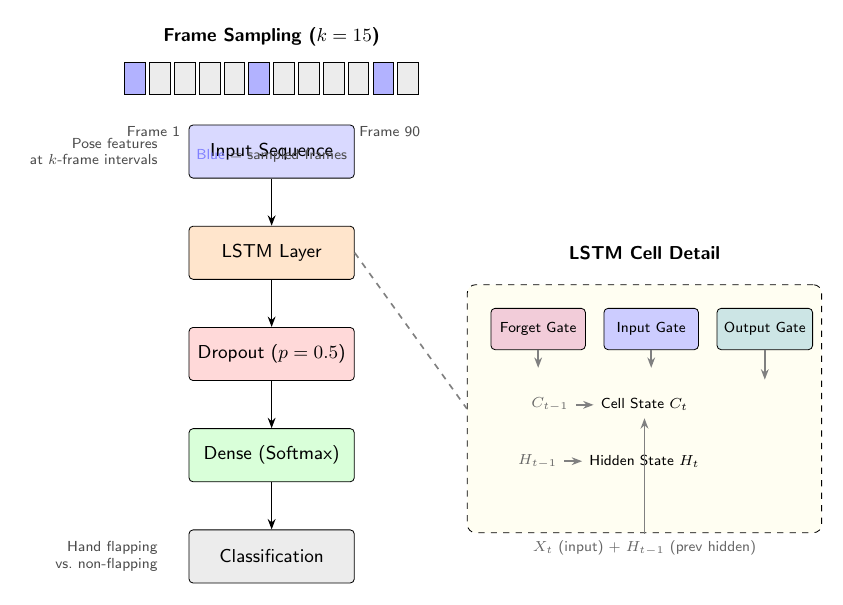}
\caption{LSTM model architecture. Input pose features extracted at $k$-frame intervals are processed by an LSTM layer, followed by dropout regularization and a dense classification layer. The LSTM cell (right) uses three gates---forget, input, and output---to selectively retain temporal information.}
\label{fig:lstm_arch}
\end{figure}

\paragraph{GRU Model.} The second model replaced the LSTM layer with a GRU layer, retaining the same dropout rate and dense output configuration. The GRU uses update and reset gates to control information flow, achieving a comparable gating mechanism with fewer parameters than the LSTM (two gates versus three), resulting in faster training~\cite{chung2014empirical}.

\begin{figure}[H]
\centering
\includegraphics[width=0.85\textwidth]{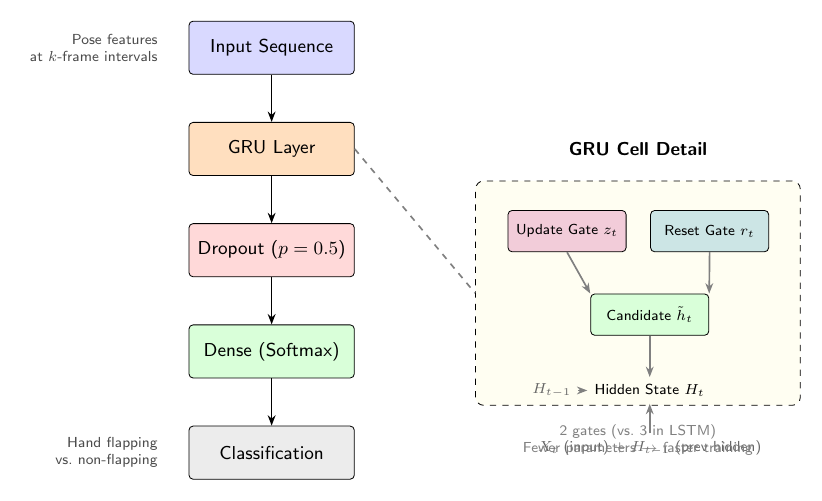}
\caption{GRU model architecture. The GRU cell uses two gates (update and reset) rather than three, reducing the parameter count and training time relative to the LSTM while maintaining comparable gating capability.}
\label{fig:gru_arch}
\end{figure}

Both models were implemented in TensorFlow/Keras and trained using categorical cross-entropy loss with the Adam optimizer. The binary classification task distinguished hand flapping from non-hand-flapping behaviors.

\subsubsection{Evaluation Protocol}

Models were evaluated using an 80-20 train-test split. Accuracy, precision, recall, and loss were recorded for each combination of architecture and sampling interval. Due to the small dataset size, measures of statistical uncertainty (e.g., confidence intervals from cross-validation) were not computed; this limitation is addressed in Section~\ref{sec:limitations}.

\subsection{Experiment 2: Data Augmentation and Transfer Learning}

\subsubsection{Preprocessing}

For the data augmentation experiments, the full SSBD dataset (75 videos, three behavior classes) was used. Two parallel feature representations were constructed: (1) RGB pixel arrays extracted from 64-frame segments of each video, and (2) optical flow vectors computed using the TV-L1 algorithm~\cite{wedel2009tvl1} on the same segments. These dual-stream representations capture both appearance and motion information.

\subsubsection{Transfer Learning with I3D}

The Inflated 3D ConvNet (I3D)~\cite{carreira2017quo}, pretrained on the Kinetics-400 dataset, served as the feature extraction backbone. The I3D input layer was extracted and incorporated as the first layer of a downstream classification network, followed by dense layers with hyperbolic tangent (tanh) and softmax activations. This transfer learning approach leverages spatiotemporal representations learned from a large-scale activity recognition corpus, mitigating the data scarcity of the SSBD dataset.

\begin{figure}[H]
\centering
\includegraphics[width=0.95\textwidth]{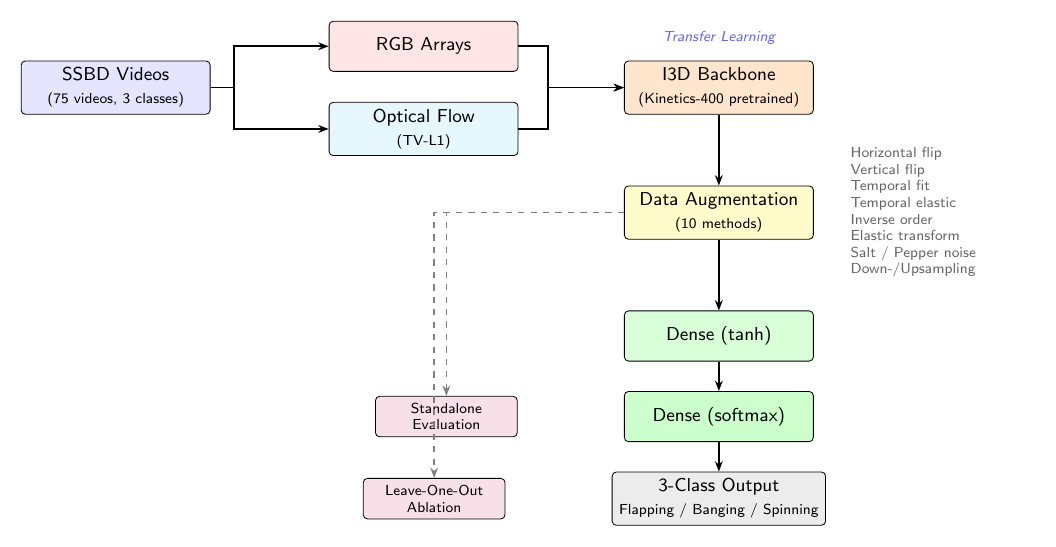}
\caption{I3D transfer learning pipeline for data augmentation experiments. SSBD videos are preprocessed into dual-stream RGB and optical flow representations, passed through the Kinetics-pretrained I3D backbone, augmented with ten strategies, and classified via dense layers. Both standalone and leave-one-out ablation evaluation protocols are applied.}
\label{fig:pipeline}
\end{figure}

\subsubsection{Data Augmentation Strategies}

Ten augmentation methods were applied to the training data:

\begin{enumerate}[nosep]
    \item \textbf{Horizontal flip}: Mirror reflection along the vertical axis.
    \item \textbf{Vertical flip}: Mirror reflection along the horizontal axis.
    \item \textbf{Temporal fit (200)}: Temporal interpolation to a fixed length of 200 frames.
    \item \textbf{Temporal elastic transformation}: Nonlinear temporal warping using random displacement fields.
    \item \textbf{Inverse order}: Reversal of the frame sequence.
    \item \textbf{Elastic transformation}: Spatial elastic deformation of individual frames.
    \item \textbf{Salt noise}: Random white pixel insertion.
    \item \textbf{Pepper noise}: Random black pixel insertion.
    \item \textbf{Downsampling}: Reduction of spatial resolution followed by re-upscaling.
    \item \textbf{Upsampling}: Interpolation to increase the number of frames in each segment.
\end{enumerate}

Two experimental protocols were used. In the \emph{standalone} protocol, each augmentation method was applied individually to measure its isolated effect on model performance (sample size: 10, number of trials: 5). In the \emph{ablation} (leave-one-out) protocol, all augmentation methods were applied simultaneously except one, and the resulting performance was compared to the all-augmentations baseline. This ablation identifies which augmentation contributes most to overall pipeline performance.

\subsubsection{Personalized Machine Learning}

A personalized machine learning protocol was also evaluated, in which a separate model was trained for each child (video) in the dataset. Each video was divided into five 16-frame segments; the first four segments served as training data and the fifth as the test set, implementing an 80-20 temporal split within each video. This per-subject approach tests whether individual behavioral patterns are sufficiently consistent within a single recording to support within-video generalization, an important consideration for clinical deployment where a model could be calibrated on an initial segment of a child's behavior.

\section{Results}

\subsection{Experiment 1: Architecture and Sampling Rate Comparison}

Table~\ref{tab:results_main} presents classification performance for LSTM and GRU models across all six sampling intervals.

\begin{table}[H]
\centering
\caption{Classification performance of LSTM and GRU models at different frame sampling intervals on the SSBD hand flapping detection task.}
\label{tab:results_main}
\begin{tabular}{llcccc}
\toprule
\textbf{Model} & \textbf{Frame Rate} & \textbf{Accuracy} & \textbf{Precision} & \textbf{Recall} & \textbf{Loss} \\
\midrule
\multirow{6}{*}{LSTM}
 & 1  & 0.900  & 0.8333 & 1.000 & 0.5715 \\
 & 5  & 0.950  & 0.950  & 0.950 & 0.1381 \\
 & 15 & \textbf{0.975} & 0.975 & 0.975 & 0.1401 \\
 & 30 & 0.950  & 0.950  & 0.950 & 0.1604 \\
 & 45 & 0.850  & 1.000  & 0.700 & 0.6009 \\
 & 90 & 0.8125 & 0.9032 & 0.700 & 0.4698 \\
\midrule
\multirow{6}{*}{GRU}
 & 1  & 0.925  & 1.000  & 0.850 & 0.2217 \\
 & 5  & 0.950  & 0.9286 & 0.975 & 0.1323 \\
 & 15 & \textbf{0.9875} & 0.8108 & 0.750 & 0.5073 \\
 & 30 & 0.8125 & 0.8205 & 0.800 & 0.4998 \\
 & 45 & 0.800  & 0.8750 & 0.700 & 0.5796 \\
 & 90 & 0.8125 & 0.9032 & 0.700 & 0.4776 \\
\bottomrule
\end{tabular}
\end{table}

\begin{figure}[H]
\centering
\includegraphics[width=0.85\textwidth]{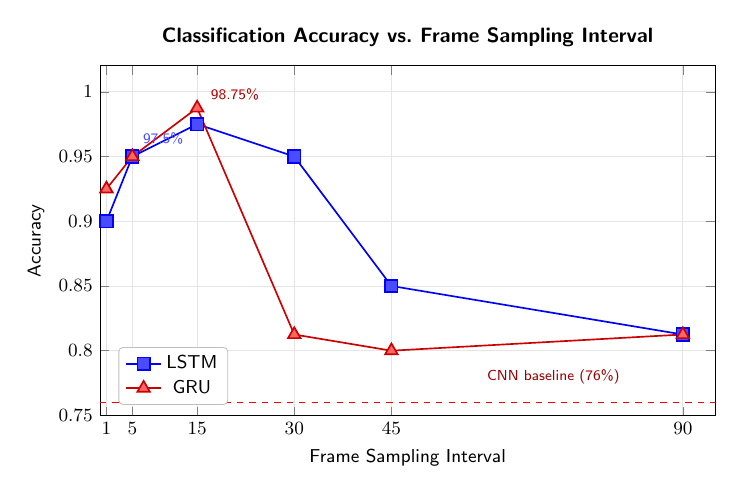}
\caption{Classification accuracy as a function of frame sampling interval for LSTM and GRU models. Both architectures peak at 15-frame intervals and substantially exceed the prior CNN baseline (dashed red line, 76\%).}
\label{fig:accuracy_plot}
\end{figure}

Both architectures achieved peak accuracy when videos were sampled every 15 frames: 97.5\% for LSTM and 98.75\% for GRU. Performance degraded at both extremes of the sampling spectrum. At a sampling interval of 1 frame (no downsampling), accuracy was 90.0\% for LSTM and 92.5\% for GRU, likely due to redundant temporal information introducing noise or overfitting. At a sampling interval of 90 frames (a single frame per video), both models produced accuracies of approximately 81\%, reflecting the loss of temporal dynamics when the input sequence collapses to a single observation.

The LSTM model produced higher average accuracy across all sampling intervals (89.6\% vs.\ 88.1\% for GRU). However, the GRU achieved the single highest accuracy value (98.75\% at 15-frame sampling), and prior work has demonstrated faster GRU training times due to its simpler gating mechanism~\cite{chung2014empirical, lakkapragada2022classification}.

Both architectures substantially outperformed the CNN-based baselines reported in Khodatars et al.~\cite{khodatars2021deep} (62--76\% accuracy), as well as the bag-of-words SVM baseline (47.1\%) reported in the original SSBD study~\cite{rajagopalan2013ssbd}.

\subsection{Experiment 2: Data Augmentation Results}

\subsubsection{Standalone Augmentation Performance}

Table~\ref{tab:aug_standalone} summarizes the three-class classification accuracy for each data augmentation method applied in isolation.

\begin{table}[H]
\centering
\caption{Standalone accuracy of individual data augmentation methods applied to the I3D transfer learning pipeline on the SSBD dataset (three-class task; sample size = 10, trials = 5).}
\label{tab:aug_standalone}
\begin{tabular}{lc}
\toprule
\textbf{Augmentation Method} & \textbf{Accuracy (\%)} \\
\midrule
Horizontal Flip              & \textbf{48.78} \\
Downsample                   & 43.90 \\
Salt Noise                   & 39.02 \\
Vertical Flip                & 34.15 \\
Elastic Transformation       & 34.15 \\
Temporal Fit (200)           & 29.27 \\
Upsample                     & 29.27 \\
Inverse Order                & 26.83 \\
Pepper Noise                 & 24.39 \\
Temporal Elastic Trans.      & 24.39 \\
\bottomrule
\end{tabular}
\end{table}

\begin{figure}[H]
\centering
\includegraphics[width=0.9\textwidth]{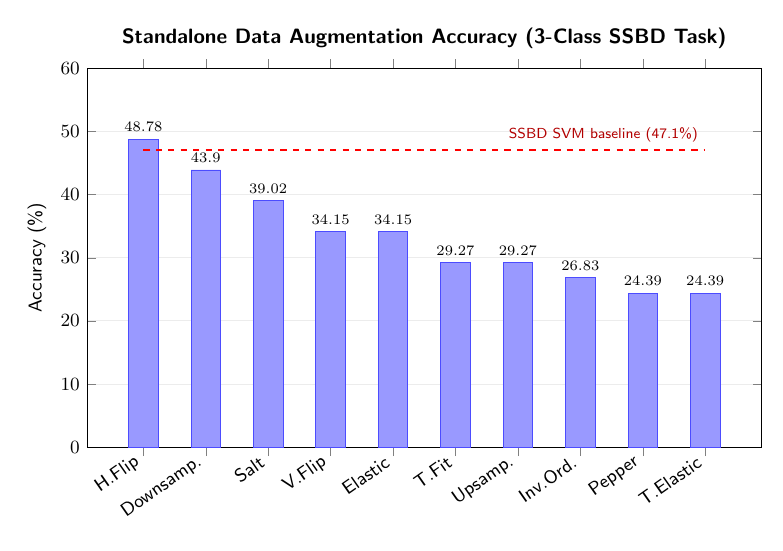}
\caption{Standalone accuracy of each data augmentation method on the three-class SSBD classification task. Spatial augmentations (horizontal flip, downsampling) outperform temporal augmentations. The dashed red line indicates the original SSBD SVM baseline (47.1\%).}
\label{fig:aug_bars}
\end{figure}

Horizontal flip was the most effective standalone augmentation (48.78\%), followed by downsampling (43.90\%). Temporal-domain augmentations (temporal fit, temporal elastic transformation, inverse order) generally underperformed spatial-domain augmentations, suggesting that spatial invariances are more beneficial than temporal perturbations for this classification task.

\subsubsection{Ablation Study}

Table~\ref{tab:aug_ablation} reports performance when each augmentation method was excluded from the full pipeline.

\begin{table}[H]
\centering
\caption{Leave-one-out ablation results. Each row shows performance when the named method is excluded from the full augmentation pipeline.}
\label{tab:aug_ablation}
\begin{tabular}{lcccc}
\toprule
\textbf{Excluded Method} & \textbf{Loss} & \textbf{Precision} & \textbf{Recall} & \textbf{Val.\ Loss} \\
\midrule
Horizontal Flip           & 3.6999 & 0.2308 & 0.2195 & 4.8482 \\
Temporal Fit (200)        & 2.6171 & 0.3158 & 0.2927 & 2.9434 \\
Temporal Elastic Trans.   & 2.0756 & 0.2973 & 0.2683 & 2.7542 \\
Inverse Order             & 2.7817 & 0.2750 & 0.2683 & 1.7527 \\
Elastic Transformation    & 3.4948 & 0.3590 & 0.3415 & 4.7026 \\
Vertical Flip             & 2.6661 & 0.3684 & 0.3415 & 1.3533 \\
Salt Noise                & 2.6837 & 0.4103 & 0.3902 & 3.1372 \\
Pepper Noise              & 2.2998 & 0.2703 & 0.2439 & 1.6560 \\
Downsample                & 1.9608 & 0.4865 & 0.4390 & 3.1856 \\
\textbf{Upsample}         & \textbf{5.0101} & 0.3000 & 0.2927 & 1.9899 \\
\bottomrule
\end{tabular}
\end{table}

Excluding upsampling produced the highest training loss (5.0101), substantially exceeding the loss observed when any other individual method was removed. This result indicates that upsampling is the single most important augmentation component for maintaining pipeline performance on this dataset. Excluding horizontal flip produced the second-highest loss (3.6999) and the lowest precision/recall among all ablation conditions, consistent with its strong standalone performance.

\subsubsection{Personalized Machine Learning}

The personalized machine learning protocol---in which a separate model was trained per child using an 80-20 temporal split within each video---produced a mean loss of 1.84 (SD = 0.79) across all subjects. The relatively low standard deviation suggests that behavioral patterns within individual videos are sufficiently consistent to support within-video transfer, although the overall loss values indicate room for improvement in absolute classification performance.

\section{Discussion}

\subsection{Architecture Selection and Sampling Rate}

The finding that both LSTM and GRU architectures substantially outperform prior CNN approaches on the SSBD hand flapping task suggests that explicit temporal modeling provides meaningful benefit for self-stimulatory behavior classification. CNNs process individual frames or short fixed-length clips, discarding the longer-range temporal structure that distinguishes repetitive stereotypies from momentary gestures. LSTM and GRU models, by contrast, maintain hidden state representations that accumulate information across the input sequence, enabling detection of the periodic, sustained motion patterns characteristic of behaviors such as hand flapping.

The consistent peak performance at a 15-frame sampling interval across both architectures is notable. At the native frame rate (every frame), models likely overfit to high-frequency noise in the pose features. At very coarse sampling intervals (45 or 90 frames), the input sequence becomes too short to capture the repetitive temporal structure that defines these behaviors. The 15-frame interval appears to strike an effective balance, preserving diagnostically relevant temporal dynamics while filtering out frame-to-frame noise. This result has practical implications for deployment: processing every 15th frame reduces computational cost by approximately 93\% relative to frame-by-frame processing, an important consideration for edge-device or mobile deployment scenarios.

The GRU model achieved the single highest accuracy (98.75\%) while offering faster training due to its reduced parameter count~\cite{chung2014empirical}. This combination of accuracy and efficiency makes GRU an attractive candidate for resource-constrained deployment contexts, including mobile health applications in low-resource settings where specialist access is limited.

\subsection{Data Augmentation Strategies}

The superior performance of spatial augmentations (horizontal flip, downsampling) over temporal augmentations (temporal elastic transformation, inverse order) is consistent with the nature of the classification task. Self-stimulatory behaviors exhibit spatial symmetry---hand flapping, for instance, occurs with comparable visual characteristics regardless of whether the child is facing left or right---making horizontal flip a natural invariance for this domain. Temporal augmentations, by contrast, may distort the periodic structure that models rely on for classification, explaining their weaker or negative contribution.

The ablation study's identification of upsampling as the most critical augmentation component is particularly informative. Upsampling increases the effective number of training frames per video, directly addressing the small dataset problem. Its removal caused the largest performance degradation, exceeding even the effect of removing horizontal flip. This suggests that for small behavioral datasets, augmentation strategies that increase the quantity of training examples may be more important than those that introduce invariance to geometric transformations.

The relatively low absolute accuracies observed in the data augmentation experiments (peak of 48.78\% on the three-class task) merit discussion. Unlike the binary classification in Experiment 1, the augmentation experiments address a three-way classification problem (arm flapping vs.\ head banging vs.\ spinning), which is inherently more challenging. The original SSBD baseline achieved only 47.1\% on this same three-class task~\cite{rajagopalan2013ssbd}, placing the augmentation results in appropriate context.

\subsection{Personalized Machine Learning}

The personalized machine learning results suggest that within-video temporal consistency is sufficient to support per-subject model training, at least to the extent of producing predictions with low variance. The practical implication is that in a clinical deployment scenario, a short initial observation period could be used to calibrate a model to a specific child's behavioral patterns before applying it to subsequent segments. This approach sidesteps some of the generalization challenges inherent in training population-level models on small datasets, though it introduces new challenges around the minimum calibration duration needed for reliable performance.

\section{Limitations}
\label{sec:limitations}

Several limitations of this study warrant discussion. First, the SSBD dataset contains only 50--75 videos depending on the experimental subset, which is insufficient to draw strong conclusions about generalizability to the broader ASD population. The absence of cross-validation or bootstrapped confidence intervals in Experiment 1 means that the reported accuracy differences between architectures and sampling rates may not be statistically significant. Future work should employ $k$-fold cross-validation with appropriate statistical tests (e.g., paired permutation tests) to establish reliability.

Second, the SSBD videos were collected from publicly posted internet recordings, introducing heterogeneity in recording conditions (camera angle, resolution, lighting, background) that may not be representative of controlled clinical or home-based screening scenarios. Some videos in the dataset were inaccessible due to preprocessing errors, reducing the effective sample size.

Third, the personalized machine learning protocol, while producing low-variance predictions, was evaluated using a single temporal split per video. The 80-20 split could be sensitive to the specific temporal boundary chosen, and the absence of multiple split evaluations limits confidence in the reported loss statistics.

Fourth, the data augmentation experiments used a transfer learning pipeline based on I3D pretrained on Kinetics-400, which was designed for general activity recognition rather than clinical behavioral analysis. Domain-specific pretraining on larger behavioral datasets, if such datasets become available, could improve downstream performance.

Finally, this study evaluated only hand-crafted sequential architectures (LSTM, GRU) with relatively simple network topologies. More recent architectures such as temporal convolutional networks (TCNs) and transformer-based models may offer superior performance and should be compared in future work.

\section{Conclusion}

This study provides two sets of empirical results relevant to automated detection of autism-related self-stimulatory behaviors. The architecture comparison demonstrates that sequence-based models (LSTM, GRU) consistently outperform prior CNN approaches for self-stimulatory behavior classification, with both architectures achieving peak performance when videos are sampled every 15 frames. The GRU architecture's combination of high accuracy (98.75\%) and lower computational cost relative to LSTM positions it as a practical choice for deployment in resource-constrained settings. The data augmentation analysis characterizes the relative effectiveness of ten augmentation strategies for small clinical behavioral video datasets, identifying horizontal flip as the strongest standalone method and upsampling as the most critical pipeline component. Together, these results equip researchers and practitioners with evidence-based guidance for architecture selection, temporal sampling, and data augmentation when building computational tools for autism behavioral screening in data-scarce clinical domains.

\bibliographystyle{unsrt}
\bibliography{references}

\end{document}